# Accurate and Energy Efficient: Local Retrieval-Augmented Generation Models Outperform Commercial Large Language Models in Medical Tasks

Konstantinos Vrettos[1], Michail E. Klontzas[#,1,2,3,5]

1. Artificial Intelligence and Translational Imaging (ATI) Lab, Department of Radiology, School of Medicine, University of Crete
2. Department of Medical Imaging, University Hospital of Heraklion, Crete, Greece
3. Division of Radiology, Department of Clinical Science Intervention and Technology (CLINTEC), Karolinska Institute, Huddinge, Sweden
4. Computational Biomedicine Laboratory, Institute of Computer Science Foundation for Research and Technology Hellas (ICS-FORTH), Heraklion, Crete, Greece

#Corresponding author

**Michail E. Klontzas, MD, PhD**

Assistant Professor of Radiology

Group Leader - Artificial Intelligence and Translational Imaging (ATI) Lab

Department of Radiology, School of Medicine, University of Crete

Voutes, 71003, Heraklion, Crete, Greece, Tel: +30 2811391351

E-mail: miklontzas@gmail.com; miklontzas@uoc.gr

ORCID: 0000-0003-2731-933X

# Abstract

Background:

The increasing adoption of Artificial Intelligence (AI) in healthcare has sparked growing concerns about its environmental and ethical implications. Commercial Large Language Models (LLMs), such as ChatGPT and DeepSeek, require substantial resources, while the utilization of these systems for medical purposes raises critical issues regarding patient privacy and safety.

Methods:

We developed a customizable Retrieval-Augmented Generation (RAG) framework for medical tasks, which monitors its energy usage and $CO_2$ emissions. This system was then used to create RAGs based on various open-source LLMs. The tested models included both general purpose models like llama3.1:8b and medgemma-4b-it, which is medical-domain specific. The best RAG's performance and energy consumption was compared to DeepSeekV3-R1 and OpenAI's o4-mini model. A dataset of medical questions was used for the evaluation.

Results:

Custom RAG models outperformed commercial models in accuracy and energy consumption. The RAG model built on llama3.1:8B achieved the highest accuracy (58.5%) and was significantly better than other models, including o4-mini and DeepSeekV3-R1. The llama3.1-RAG also exhibited the lowest energy consumption and $CO_2$ footprint among all models, with a Performance per kWh of 0.52 and a total $CO_2$ emission of 473g. Compared to o4-mini, the llama3.1-RAG achieved 2.7x times more accuracy points per kWh and 172% less electricity usage while maintaining higher accuracy.

Conclusion:

Our study demonstrates that local LLMs can be leveraged to develop RAGs that outperform commercial, online LLMs in medical tasks, while having a smaller environmental impact. Our modular framework promotes sustainable AI development, reducing electricity usage and aligning with the UN's Sustainable Development Goals.As AI systems

**Description:** A customizable RAG framework for medical tasks is presented. Use of the framework indicates that local custom LLMs outperform commercial LLMs in accuracy and energy efficiency for medical tasks

# Introduction

As AI systems are increasingly utilized in healthcare, the environmental[1] and ethical challenges[2] associated with their operation have become paramount concerns. Employing Large language models (LLMs), such as ChatGPT and DeepSeek, while powerful tools for generating responses, requires vast amounts of resources[3]. The cluster centers that host these models contribute substantially to global environmental degradation through increased water consumption, electricity usage, and reliance on rare earth minerals[4]. Reports suggest that the OpenAI clusters used up to 6% of a districts water and consumed as much electricity as 33000 homes[5].

The generation of medical information by these systems raises further concerns, including potential biases, lack of transparency about source credibility and risks of misinformation or harmful answers[6]. The use of online LLMs in medicine also raises critical privacy issues. Medical data, when sourced to external platforms, may expose sensitive patient information to unauthorized entities, compromising confidentiality and raising serious ethical issues. On the other hand, it has also been shown that the default versions of open-access, local LLMs do not perform well on medical tasks[7]. Consequently, a more robust framework is needed for secure and controlled access to medical knowledge within healthcare environments.

These challenges can be addressed by using a Retrieval-Augmented Generation (RAG) model specifically tailored for medical applications. First, it ensures enhanced data privacy and security by keeping sensitive medical information within the organization's secure environment, complying with regulations like HIPAA or GDPR. This system also allows tailored retrieval from vetted medical sources, which avoids inaccuracies and has been shown to outperform general-purpose models[8]. Auditability is another critical feature, as it allows clinicians to verify the sources of answers, increasing trust in the outputs. To our knowledge there have been no prior efforts to use local LLMs for medical RAGs in an environmentally-conscious way.

Our primary objective is to develop an open-source modular framework which will enable users to build customized medical RAGs while controlling its energy usage. To prove the capabilities and energy efficiency of this approach, we aim to create a local, medical RAG and compare its accuracy and energy efficiency to that of closed-source LLMs like o4-mini(OpenAI). The use of local LLMs can enables the containment of Personal Identifiable Information (PII) within the hospital network, alleviating privacy concerns. Another objective is to develop a RAG methodology that operates efficiently within the constraints of consumer-grade hardware, thus ensuring that our model is both effective and accessible. This approach supports the broader goal of advancing healthcare in an environmentally responsible manner[9]. The proposed RAG method seeks to balance the growing demand for intelligent medical tools with the urgent need to mitigate their environmental impact and address current limitations of online LLMs.

## Methods

Dataset

The MedQA [10] and the PubMedQA [11] datasets were used to evaluate the ability of RAG models to provide accurate and relevant zero-shot answers within a clinical context. The dataset used herein included 1,000 multiple-choice questions with answers, divided equally into two categories: 500 questions focused on patient cases and 500 questions related to advances in medical research. A total of 500 English patient case questions were randomly selected ("random" module in python) from the MedQA dataset, to assess the capability of the RAG models to answer clinical-related questions. Questions were equally split between USMLE "step 1" and "step 2" levels, ensuring a diverse range of clinical scenarios and topics. Each question was paired with four or more multiple choice answers. This benchmark has been widely used to test the performance of LLMs in medical domain. To assess the capability of the models to answer questions on emerging research advancements, 500 questions regarding scientific advances were randomly selected ("random" module in python) from the PubMedQA dataset[11]. This dataset contains questions on biomedical research articles from PubMed which test the RAG model's ability to reason scientifically. All PubMedQA questions were stripped from their expert-selected context, leaving only the question and answers.

Retrieval Augmented Generation framework

The RAG framework (Figure 1) was designed, using Python (3.9v), to enhance the performance of a local language model on domain-specific tasks, particularly focusing on medical applications. First, the curated medical literature is organized into chunks and indexed using FAISS (1.5.3v), a library that facilitates efficient approximate nearest neighbor searches[12]. The data is then transformed into a high-dimensional vector representation using OllamaEmbeddings (langchain-community 0.2.16v) with the "mxbai-embed-large" model. We used the "Cecil Textbook of Medicine"[13] as corpus. During inference, the user query is used by the Retriever module to return the relevant documents from the structured corpus. Then a prompt that contains the user's query and the retrieved data is constructed. This prompt is fed to the Generator component which creates responses through a local language model. The electricity usage and $CO_2$ footprint of the entire pipeline are monitored using the carbontracker package (2.2.0v) and presented together with the answer of the model. All processing is conducted locally, on a system with 64GB of RAM and an NVIDIA RTX 5000 with 32GB of VRAM. This framework is available on github: https://github.com/konstvr/GreenRAG

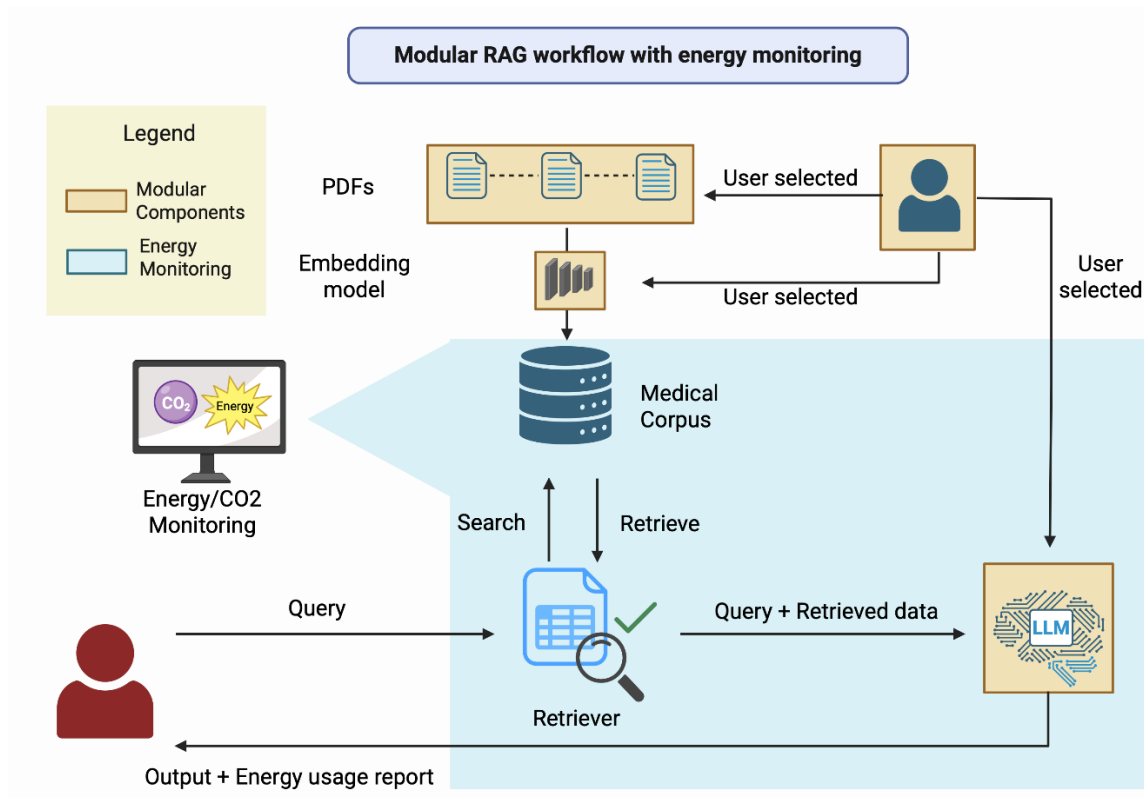

Figure 1: Modular RAG framework, with energy usage and $CO_2$ monitoring. The elements in yellow can be customised by the user.

LLM Selection and RAG Evaluation

In order to select which open access model will be used as the basis for the Generator component of the RAG, five LLMs were evaluated. The assessment includes both general purpose models: llama3:8B, llama3.1:8B, llama3.2:1B, mistral:7B and medgemma-4b-it, which is the latest medical-domain model released by Google. These models have been selected because they have the best reported performance[14] while being under 8 billion parameters, which guarantees functionality on consumer-grade hardware without the need for High Performance Computing resources. This can enable local processing for the protection of PII. RAG was constructed using each of these models and its performance was tested on the questions dataset, while the consumed energy in each case was monitored. The performance of OpenAI's o4-mini (not to be confused with ChatGPT), which is optimized for cost-efficient reasoning, and DeepSeekV3-R1 was also measured on our dataset, as a reference. To compute the energy consumption and $CO_2$ output of these closed-source models we relied on existing reports, as it is not possible to directly calculate the energy usage of the online models. For DeepSeek and o4-mini the "Web Search" was activated, to ensure up-to-date responses. For all evaluated models the following metrics were used: accuracy, precision, recall, f1 score, latency per question, throughput (Q/sec), energy consumption (CPU and GPU) in kWh, Performance per kWh ($PPW_{kWh}$) and $CO_2$ footprint in grams. $PPW_{kWh}$ is calculated as: Accuracy(avg) / Total Energy(kWh). $CO_2$ footprint (in grams) is calculated as: kWh * Carbon Intensity ($gCO_2$/kWh), where Carbon Intensity is 0.43 for local LLMs (Greece), 0.65 for DeepSeek (China) and 0.38 for o4-mini (Germany). The models' performance was compared based on accuracy using Wilcoxon's signed rank test and the threshold for statistical significance is $p<0.05$.

# Results

Performance analysis

The best performing model was the RAG built on llama3.1:8B (llama3.1-RAG), which achieved an accuracy of 58.5% (55.44%, 61.55%) and f1-score of 57% (53%,60%). The llama3.1-RAG was significantly better, in terms of accuracy, than the RAGs that were based on llama3.2:1B, mistral:7B, medgemma-4b-it ($p<0.05$ for all comparisons) and non-inferior to the llama3-RAG ($p=0.06$). The online LLMs, GPT-4 and DeepSeekV3-R1 scored an accuracy of 57% (50.14%, 63.86%) and 47.5% (40.58%, 54.42%) respectively. The llama3.1-RAG performed significantly better than o4-mini and Deepseek (p values<0.05). For both o4-mini and DeepSeek the "Web Search" option was activated. Highest throughput and thus lower latency were achieved by o4-mini and DeepSeek at 0.5 sec/Q and 2 Q/sec. In addition, we found that prompt engineering improved the performance of the RAG by 9% for MedGemma-RAG and 4% for llama3.1-RAG. The performance of all evaluated models is summarized in Table 1.

**Table 1**: Performance, Energy usage, CO$_2$ emissions per model

| Model Name | Latency per Q (sec) | Throughput Q/sec | Accuracy | Precision (avg) | Recall (avg) | F1 (avg) | Energy consumption (GPU)(kWH) | Energy consumption (CPU)(kWH) | Energy consumption (TOTAL)(kWH) | CO2 footprint(grams) | PPW(kWh) |
|---|---|---|---|---|---|---|---|---|---|---|---|
| **RAG-llama3.1:8b** | 2.99 | 0.33 | 58.5% [55.44%, 61.55%] | 58% | 57% | 57% [53%,60%] | 0.07 | 1 | 1.1 | 473 | 0.53 |
| **RAG-MedGemma:4b** | 6.79 | 0.15 | 47.5% [44.4%, 50.6%] | 51% | 47% | 42% [38%,45%] | 0.13 | 2.29 | 2.46 | 1057.8 | 0.19 |
| **RAG-llama3.2:1b** | 2.5 | 0.4 | 32.97% [30.05%, 35.88%] | 37% | 33% | 32.73% [29.6%, 35.76%] | 0.03 | 0.87 | 0.92 | 395.6 | 0.35 |
| **RAG-mistral:7b** | 3.11 | 0.32 | 27.1% [24.35%, 29.86%] | 45% | 27% | 25.48% [22.55%, 28.33%] | 0.08 | 1.07 | 1.17 | 503.1 | 0.23 |
| **RAG-llama3:8b** | 3.12 | 0.32 | 55.96% [52.88%, 59.03%] | 56% | 56% | 53.86% [50.57%, 57.05%] | 0.07 | 1 | 1.1 | 473 | 0.50 |
| **o4-mini-Search** | 0.5 | 2 | 57% [50.14%, 63.86%] | 52% | 62% | 56.8% [48.54%, 53.66%] | - | - | 3 | 1140 | 0.19 |
| **DeepSeekV3-R1-Search** | 0.5 | 2 | 47.5% [40.58%, 54.42%] | 52% | 62% | 46.99% [37.41%, 43.99%] | - | - | 3 | 1950 | 0.16 |

Energy efficiency

The environmental impact and energy efficiency of each model were compared by measuring $CO_2$ footprint (Figure 2) and calculating the PPW score (Figure 3). The most efficient model was Llama3.1-RAG which achieved a $PPW_{kWh}$ of 0.52 and an emitted 473g of $CO_2$ in total. The rest open-access LLMs had energy consumption and $CO_2$ emissions similar to llama3.1-RAG except for MedGemma-RAG, which had a $PPW_{kWh}$ of 0.19 while emitting 1057.8g of $CO_2$. Total CPU/GPU electricity usage in kWh was 0.07/1 for llama3.1-RAG and 0.13/2.29 for MedGemma-RAG. The estimation of energy consumption and $CO_2$ emissions for GPT and DeepSeek was based on the widely accepted assumption that a simple GPT-4 prompt uses 3Wh [15]. Regarding online DeepSeekV3-R1, the energy consumption per prompt has been calculated for the 70B parameter version[16]. The report shows that the 70B DeepSeek version generates 0.07 tokens per watt-second and an average answer to a question in our dataset is 900 tokens (including reasoning). As a result, the 70B version of DeepSeek would consume 3.57Wh per prompt. The online version of DeepSeek is 671B parameters and it has been shown that larger LLMs consume more energy[17]. Based on this evidence, we estimated that the closed-source models use 3 Wh per prompt, which is the most optimistic number since we also had the "Web Search" option turned on, for both models. Prior works have estimated o4-mini and DeepSeekV3 could be using up to 5.1Wh and 9.1Wh respectively, for a medium sized answer[18]. Consequently, 04-mini had a PPW 0.19 and a $CO_2$ footprint 1140g of, while DeepSeek had a PPW of 0.15 and produced 1950g of $CO_2$. Accuracy and $CO_2$ emissions for all models are summarized in Table 1 and Figure 3.

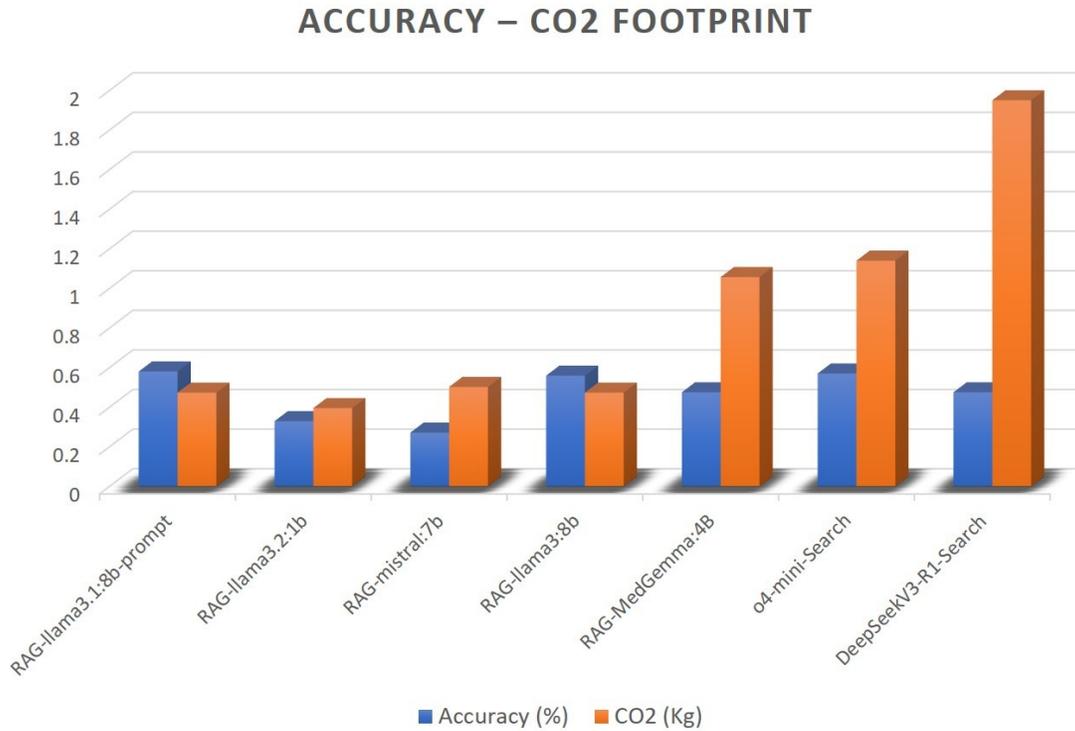

Figure 2: Comparison of accuracy (%) and $CO_2$ emissions among all models

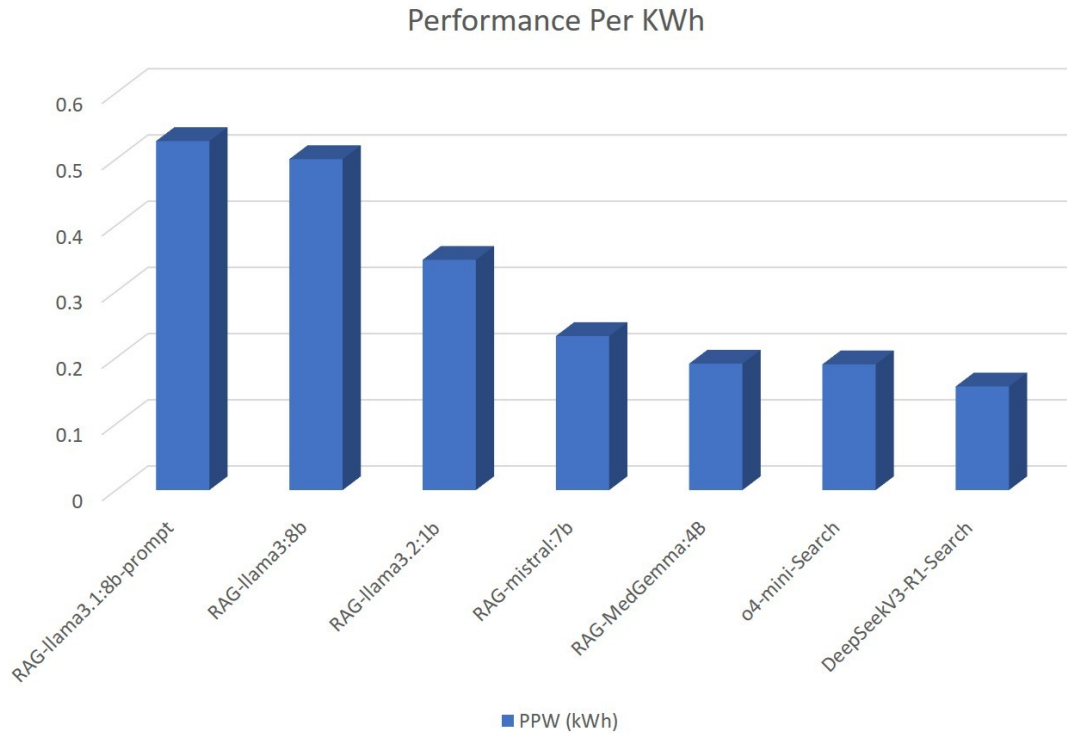

Figure 3: Comparison of Performance Per KWh among all models

# Discussion

The development of accurate and sustainable medical language models has become increasingly important in recent years. This study aimed to address this challenge by proposing a modular RAG framework based on local LLMs that monitors its energy usage and $CO_2$ output. The modular design of our framework enables users to easily modify and update the corpus of the RAG, without having to start from scratch. This flexibility is particularly important in the medical domain, where new research and discoveries are constantly being made. By allowing users to develop their own local RAGs, we promote sustainable AI development, while maintaining privacy and considering the environmental impact of AI systems.

We demonstrated the efficacy of our proposed framework by using it to develop a RAG for medical question answering, which outperformed commercial/online LLMs like o4-mini while being more environmentally friendly. The llama3.1-RAG achieved 2.7x times more accuracy points per kWh and 172% less electricity usage than o4-mini while achieving higher accuracy. This is consistent with previous work by Samsi et al, who showed that a large LLM (67B) uses significantly more energy than a model of 7B parameters[17]. The performance of the RAG was influenced by several factors, including prompt engineering and which local LLM was chosen as the basis for the RAG. An intriguing finding is that prompt engineering has a more significant impact on the performance of the RAG than using a specialized medical domain LLM, like the recently released MedGemma. A similar conclusion was reached by Rubei et al, who showed that prompt engineering can improve the LLM's accuracy while also reducing energy usage[19]. In addition, functionality and accessibility have been prioritized in the development of our RAG, which means that it can be used on consumer-grade hardware. A previous work by Xiong et al, evaluated several RAG models on multiple-choice medical questions and their best open-access model (Mixtral 4*7B parameters) achieved an accuracy of 60%[20], which is similar to our model's 58.5%. However, the LLM used in our case was much smaller (8B parameters) and thus can be more easily implemented by the community.

The health sector is one of the main emitters of $CO_2$[21] and the growing environmental impact of large-scale AI systems has become a pressing concern in recent years. Consequently, it is essential to start addressing this issue by incorporating the environmental cost, into the development of AI tools for healthcare and prioritizing efficient models[22]. Our proposed RAG framework is a step towards achieving this goal, as it allows users to monitor the energy usage and $CO_2$ output of their local RAGs. This aligns with the United Nations Sustainable Development Goals (SDGs), which aim to ensure sustainable consumption and production patterns. Our proposed framework addresses this issue by allowing users to develop and train their own local RAGs, which can be tailored to their needs, while keeping track of the environmental impact of their solution.

While the results demonstrate the efficacy of the proposed RAG framework, there are several limitations to this study that need to be acknowledged. Firstly, the performance of the RAG was not evaluated on open-ended questions. Secondly, while the RAG achieved higher accuracy than online LLMs, using a wider corpus, optimizing the prompts or using larger LLMs could potentially yield better performance. We avoided using models bigger than 8B parameters in order to reduce electricity usage and make the model accessible to users with low-end hardware. These limitations highlight areas for future research and development.

## Conclusion

In conclusion, this study demonstrates that local LLMs can be leveraged to develop RAGs that not only outperform commercial online LLMs in medical question answering tasks but also offer significant environmental benefits. By designing and releasing a modular framework that enables users to monitor and optimize their RAG's energy usage and $CO_2$ output, we aim to promote sustainable AI development in the healthcare sector. Our results show that local RAGs can achieve higher accuracy while reducing electricity usage by up to 172% compared to online LLMs, making them an attractive solution for medical professionals seeking efficient and environmentally friendly tools. As the health sector continues to grapple with its environmental footprint, our proposed framework offers a practical step towards ensuring that AI systems are developed in a responsible and sustainable manner.